# BdSLW60: A Word-Level Bangla Sign Language Dataset

Husne Ara Rubaiyeat[1], Hasan Mahmud[2*], Ahsan Habib[2], Md. Kamrul Hasan[2]

[1]Computer Science Department, Natural Science Group, National University, Bangladesh, Board Bazar, Gazipur, 1704, Dhaka, Bangladesh.
[2]Systems and Software Lab (SSL), Department of Computer Science and Engineering (CSE), Islamic University of Technology (IUT), OIC, Board Bazar, Gazipur, 1704, Dhaka, Bangladesh.

*Corresponding author(s). E-mail(s): hasan@iut-dhaka.edu;
Contributing authors: rubaiyeat@yahoo.com;
ahsanhabib3@iut-dhaka.edu; hasank@iut-dhaka.edu;

**Abstract**

Sign language discourse is an essential mode of daily communication for the deaf and hard-of-hearing people. However, research on Bangla Sign Language (BdSL) faces notable limitations, primarily due to the lack of datasets. Recognizing word-level signs in BdSL (WL-BdSL) presents a multitude of challenges, including the need for well-annotated datasets, capturing the dynamic nature of sign gestures from facial or hand landmarks, developing suitable machine learning or deep learning-based models with substantial video samples, and so on. In this paper, we address these challenges by creating a comprehensive BdSL word-level dataset named as BdSLW60 in an un-constrained and natural setting, allowing positional and temporal variations and allowing sign users to freely change hand dominance. The dataset encompasses a total of 60 Bangla sign words, with a significant scale of 9307 video trials provided by 18 signers under the supervision of sign language professional. The dataset was rigorously annotated and cross-checked by 60 annotators. We also introduced a unique approach of a relative quantization-based key frame encoding technique for landmark based sign gesture recognition. We report the benchmarking of our BdSLW60 dataset using the Support Vector Machine (SVM) with testing accuracy up to 67.6% and an attention-based bi-LSTM with testing accuracy up to 75.1%. The dataset is available at https://www.kaggle.com/datasets/hasaniut/bdslw60 and the code base is accessible from https://github.com/hasanssl/BdSLW60_Code.

# 1 Introduction

Sign Language (SL) is the day-to-day communication mechanism of Deaf and hard-of-hearing people. However, their communication with hearing people is still facing a lot of barriers because regular people do not have enough vocabulary in the language. Even though it is not hard to learn them, remembering a few thousand-word signs of sign language can be daunting and only feasible for sign language experts. Continuous sign Language recognition by video analysis and Machine Learning (ML) or Deep Learning (DL) techniques could be a solution to the alleviation of this problem [1].

Continuous sign language recognition means translating sign sentences to natural language. Each sign sentence is normally constructed with a few sign words, which are the building blocks of the language. In cases where there is no sign word defined for a word, or to spell a proper noun, fingerspelling is used. Recognition of sign words is the intermediary step to continuous sign language recognition[1, 2].

There are two subtle problems in Sign language research. Firstly, the scarcity of proper datasets, and secondly, the suitable ML/DL-based models that can capture the articulated sign context. The most researched sign language is American Sign Language (ASL). Researchers of ASL have recently illustrated the need for large-scale word-level sign language recognition [1, 3, 4], and finding out ways for continuous ASL recognition [1, 2, 5–7]. This stimulated researches in others languages vigorously[8–15].

There are more than 5000 sign words in BdSL[16]. A Standard effort was made by Centre for Disability in Development (CDD)[17] by video documenting those 5000 words on Compact Disks (CD). Recently in the year 2020 in an effort to popularize BdSL, CDD published mobile apps[18–23]. Still We found very limited work in the WL-BdSL dataset, specifically mentioning, only [24, 25]. In [25], the authors presented 11 daily used sign words consisting of 1105 sign images. They worked on static images, restricting their effort on WL-BdSL only to those sign words that do not need video. They reported the accuracy comparisons of different state-of-the-art transfer learning-based approaches like VGG16, VGG19, InceptionV3, and Alexnet [26]. The dataset they used consisted of static images of sign words which in no way represent continuous WL-BdSL. The models they used are mostly suitable for image instance based classification, not for spatio-temporal data-driven WL-BdSL classification like SVM-DTW [27, 28], bi-LSTM[29], 3DCNN[30], and so on.

In [24], the researchers have worked with 100 frequently used Bangla sign words but the dataset is not publicly available. Their work, based on their previous publications [31, 32], is concentrated on time-efficient feature construction for real-time WL-BdSL recognition. They have used hand crafted features to model a subset of sign linguistic features. The approach has limitations in capturing the full dynamism in local (e.g. finger movements) and global sign gestures (e.g. lateral hand movements).

The state-of-the-art methodology suggests automated features are better for learning in the recent deep-learning based advancements [4, 9, 29, 30] unless we define specific sophisticated fine gestures for specific HCI tasks. Obviously hand crafted features are not fit for classification of large number of sign word recognition challenge e.g. [4, 9, 10, 15]. Here the challenges are in the domain of sign language communication, SL translations, sign vocabulary generation, not for specific adaptive gestures to interact where we need handcrafted features. After this comprehensive study we can enumerate specific research challenges in WL-BdSL domain as follows:

- Scarcity of WL-BdSL dataset constructed in unconstrained environment verified by Bangla sign language professional and well-annotated in authentic settings.
- The need to develop appropriate ML/DL models specially designed to recognize BdSL. The model should be robust on left/right hand dominance, pose variations, temporal variations, capturing sign nuances, and so on.
- BdSL has local fine-grained gestural movements (e.g. finger, palm, wrist point movements) along with coarse-grained information (e.g. head, shoulder, arm points movements) that needs to be captured and highlighted to understand the contextual meaning. For example, hand landmark points have more sign gestural movement information than facial landmarks. However, the keypoint-based approaches may suffer from contrast variations in coordinates which needs to be addressed.

We feel that a dataset of WL-BdSL, facilitating ML/DL research is the most important work for BdSL research community. In this paper, we present BdSLW60 a comprehensive dataset of 60 words of BdSL. We also benchmark the BdSLW60 using Google media pipe-based landmarks[33] as inputs to ML/DL models. The choice of landmark-based recognition is driven by its lightness compared to vision-based approaches like 3DNNN, I3D[4, 10] and our expertise in landmark-based techniques. The landmarks are 3D points consisting of x, y, and d of pose, face, and hands.

The dataset will open the scope for immediate research in WL-BdSL. While doing experiments we realized the need for highlighting the local changes to the ML/DL models and came up with a novel idea of relative-quantization-based encoding of keyframes. We assume this is our third contribution as the idea may open the scope of directly feeding keyframes to NLP-based Large Language Models (LLM) [7, 34] or to map gesture to a lexical sign notation: Hamburg Notation System(HamNoSys) [35].

So, to enumerate, the key contributions of this paper are as follows:

- Construction of a comprehensive WL-BdSL dataset, named as BdSLW60.
- Benchmarking the performance of the recognition models on BdSLW60 using classical ML (SVM) and DL (attention-based bi-LSTM) in WL-BdSL with the landmark-based features provided by Google MediaPipe.
- Proposing a relative quantization-based key frame encoding approach for continuous sign language recognition.

## 2 Related Works

In this section, we review the BdSL datasets (Table 1) consisting of static and dynamic Bangla hand signs, their modality, availability, sign language recognition models [24, 25, 36–46], the necessity of authentic and reliable WL-BdSL dataset, and research opportunities in continuous BdSL translation using language-based models used in other sign languages [2, 6, 7, 34].

Sign languages have two types of gestures, first, static gestures representing alphabets, digits, and occasionally some word gestures represented in static images, and second, dynamic gestures for words and sentences. Sentence and word-level continuous signs may include soundless mouth utterances along with facial expressions.

Fingerspelling in sign communication is a process of controlled finger motion with specific handshapes for each constituting letter of the word to be spelled. While fingerspelling any word is very slow for communication, normally there is a dynamic gesture defined for a word in SL. If the sign for a word is expressed using words from the spoken language then it is a sign gloss. The video representation of a word or phrase in sign language is called a 'gloss' [1, 4]. A combination of glossed signs makes a sentence for communication. Sign language translation is thought of as two-stage modeling: video2gloss and gloss2text[1], though recent researches are exploring end-to-end continuous sign language translation [2, 6, 13]. Word-level sign language datasets can be used in video2gloss recognition. However, word-level sign language recognition means classifying videos to sign word classes [4, 47]. There have been numerous works on word-level sign language datasets that contain glosses/words, video modality along with both RGB and depth information for different sign languages like Turkish, Argentinian, Polish, Chinese, German, Spanish, Russian and so on [1, 4, 8, 9, 14, 15]. Most of them previously suffered from either insufficient sign instances or a lack of authentic unconstrained settings during data collection. To capture the inherent nuances and difficulties of signs we require a large dataset in natural settings often guided by sign professionals. In this regard, we found very limited work BdSL, specifically on WL-BdSL.

BdSL research started with the release of Ishara-Lipi [36] and Ishara-Rochon[37]. The two datasets consist of sign images of Bangla alphabets and digits. However, it took a while to get momentum even in this static BdSL research, and till now research articles are being published on static BdSL [45]. Recent static BdSL recognition papers [39–43, 45, 46] used Image-based approaches and standard deep learning techniques like YOLO,CNN, Alexnet, Densenet, Resnet, Squeezenet, and VGG19. Works [38, 44] used MediaPipe landmark-based features and SVM, and ANN as classifiers respectively. Recently a work [48] in the Assamese language (closely related to Bangla) also used a landmark-based feature in ANN for recognition of 9 vowels and consonants.

WL-BDSL is still in its infancy. Work [25] collected 1105 image samples of 11 static Bangla sign words and used different transfer learning techniques like VGG16, VGG19, AlexNet, and InceptionV3 with pre-trained weights [26]. This is the first work on WL-BdSL. However, the gestures they picked are static, and hence the data and classification techniques cannot be used in video-based continuous WL-BdSL. Hasanuzzaman and his team have been working on continuous BdSL recognition for a long time. Their works are concentrated on finger spelling and real-time recognition [31, 32]. Interestingly, one of their work has recently been published on WL-BdSL [24] where they used image processing-based techniques to model a subset of sign linguistic features such as motion, hand shape, and position. Due to their focus on real-time recognition, they avoid a deep learning-based approach. We found a few limitations of this work: if sign linguistic features are to be derived, all of them should be modeled to recognize a huge number of word classes, and if possible should be converted to a standard cross-linguistic sign notation like HamNoSys. In this particular work, they modeled a subset of sign linguistic features and even did not capture the motion of both hands that many sign words will require. They also introduced three weight coefficients for the three linguistic features, values of which are subjective and experimental. Moreover, the dataset they gathered and worked on is not publicly available.

**Table 1**: BdSL Datasets

| Dataset | Type | Video/Image | # of Samples | Year | Public |
| --- | --- | --- | --- | --- | --- |
| Ishara-Lipi [36] | Alphabets | Image | 1800 | 2018 | Yes |
| Ishara-Bochon [37] | Digits | Image | 1000 | 2018 | Yes |
| Rayeed et.al. [38] | Digits | Image | 10,000 | 2021 | No |
| KU-BdSL [39] | Alphabets | Image | 1500 | 2021 | Yes |
| BdSL36 [40] | Alphabets | Image | 40,000 | 2021 | Yes |
| OkkhorNama [41] | Alphanumeric | Image | 12,000+ | 2021 | Yes |
| BdSL-D1500 [42] | Alphanumeric | Image | 132,061 | 2022 | Yes |
| BdSL49 [43] | Alphanumeric | Image | 29,490 | 2023 | Yes |
| BdSL47 [44] | Alphanumeric | Image | 47,000 | 2023 | Yes |
| Siddique et.al. [46] | Alphanumeric | Image | 3,760 | 2023 | No |
| BdSL_OPSA22_STATIC1 [45] | Alphanumeric | Image | 24,615 | 2024 | Yes |
| BdSL_OPSA22_STATIC2 [45] | Alphanumeric | Image | 8,437 | 2024 | Yes |
| BdSLW-11 [25] | 11 static Words | Image | 1,105 | 2022 | Yes |
| Rahaman et.al. [24] | 100 Words | Video |  | 2023 | No |
| BdSLW60 (Our dataset) | 60 Words | Video | 9307 | 2024 | Yes |

With the advancement of DL models such as LSTM, BERT, BiLSTM, TGCN and LLM, recently there has been a research trend established in solving sign language recognition problem [1, 34, 49]. And with the advent of MediaPipe holistic[33] based landmark extraction technique, many researches are spawning out in sign language recognition domain [11, 47–51].

In this paper, we present BdSLW60, a dataset of 60 words with 9307 samples signed by 18 sign users, verified by sign experts, and duly annotated by 60 annotators. We take MediaPipe landmarks instead of a computer vision-based approach. We establish benchmarks using SVM, SVM-DTW, and Attention-based Bi-LSTM. Further, we highlight the local features of hand, pose, and face by relative quantization that produce better accuracy. Inspired by the effectiveness of a fixed quantization approach in [29] for dynamic hand gesture recognition task, we see the prospect of a novel relative quantization-based key frame encoding technique to be used as the encoding of frames in end-to-end transformers [2, 6, 50] and LLMs [34].

# 3 BdSLW60 dataset

## 3.1 Dataset Collection process

Our WL-BdSL dataset was collected in an unconstrained setting, where the volunteers participated in a Bangla Sign Language Workshop conducted by a sign language professional. The workshop spanned 8 days with additional make-up sessions provided for volunteers who missed a session and needed to catch up. Adequate support people (e.g. administering the session by providing logistics, technical support, etc.) were engaged as required. Both the expert/sign language professional and supporting individuals were remunerated appropriately.

We made a campaign to attract volunteers to join the workshop emphasizing the flexibility for participants to learn freely and leave without mandatory video recording of the sessions. The volunteer who willingly consented to contribute data, recorded their sessions themselves after being fully informed about the purpose and providing explicit consent. Those who finished all the sessions were rewarded with a token of appreciation and a certificate. Though we have collected 401 words in two camera views during the campaign but due to insufficient annotation, we are reporting only 60 Bangla sign words (front view) here (in Table 2).

Table 2: BdSLW60 Dataset Description

| Class Index | Label | Sign Word | Pronunciation | English Meaning |
|---|---|---|---|---|
| 0 | W1 | বাবা/আব্বা | bābā/ābbā | Father |
| 1 | W2 | আত্মীয় | ātmī a | Relative |
| 2 | W3 | ভাই | bhā'i | Brother |
| 3 | W4 | বোন | bōna | Sister |
| 4 | W5 | বউ | ba'u | Wife |
| 5 | W6 | চাচা | cācā | Paternal Uncle |
| 6 | W7 | চাচী | cācī | Paternal Aunt |
| 7 | W8 | দাদা/নানা | dādā/nānā | Grandfather |
| 8 | W9 | দাদী/নানী | dādī/nānī | Grandmother |
| 9 | W10 | দায়িত্ব | dā itba | Responsibility |
| 10 | W11 | দেবর | dēbar | Husband's Younger Brother |
| 11 | W12 | দুলাভাই | dulābhā'i | Sister's Husband |
| 12 | W19 | কন্যা | kan'yā | Daughter |
| 13 | W20 | মা | mā | Mother |
| 14 | W37 | আম | āma | Mango |
| 15 | W38 | আলু | ālu | Potato |
| 16 | W39 | আনারস | ānārasa | Pineapple |
| 17 | W40 | আঙ্গুর | āṅgura | Grapes |
| 18 | W41 | আপেল | āpēla | Apple |
| 19 | W42 | বিস্কুট | biskuṭa | Biscuits |
| 20 | W43 | বরই | bara'i | Jujube/Chinese Date |
| 21 | W44 | কেক | kēka | Cake |
| 22 | W45 | চা | cā | Tea |

| Continuation of Table 2 ||||| 
| Class Index | Label | Sign Word | Pronunciation | English Meaning |
| --- | --- | --- | --- | --- |
| 23 | W46 | চাল | cāla | Rice |
| 24 | W47 | চিনি | cini | Sugar |
| 25 | W48 | চিপস | cipasa | Chips |
| 26 | W49 | চকলেট | cakalēṭa | Chocolate |
| 27 | W50 | ডাল | dāla | Lentils |
| 28 | W91 | বোতাম | bōtāma | Button |
| 29 | W92 | টুপি | ṭupi | Cap |
| 30 | W93 | চাদর | cādara | shawl |
| 31 | W94 | চিরুনি | ciruni | Comb |
| 32 | W95 | চশমা | caśamā | Spectacles |
| 33 | W96 | চুড়ি | cuṛi | Bangles |
| 34 | W97 | ক্লিপ | klipa | Clip |
| 35 | W98 | ক্রিম | krima | Cream |
| 36 | W99 | তথ্য | tathya | Data |
| 37 | W100 | দেনাদার | dēnādāra | Indebted |
| 38 | W111 | জমজ | jamaja | Twin baby |
| 39 | W112 | জুতা | jutā | Shoe |
| 40 | W211 | টুথপেস্ট | tuthapēsṭa | Toothpaste |
| 41 | W212 | টিশার্ট | tiśārṭa | Tshirt |
| 42 | W213 | টিউবলাইট | ti'ubalā'iṭa | Tubelight |
| 43 | W214 | টিভি | Ṭibhi | television |
| 44 | W215 | এসি | ēsi | Air-conditioner |
| 45 | W216 | অ্যাপার্টমেন্ট | ayāpārṭamēnṭa | Apartment |
| 46 | W217 | অডিও ক্যাসেট | aḍi'ō kyāsēṭa | Audio Cassette |
| 47 | W218 | আয়না | ā anā | Looking Mirror |
| 48 | W219 | বালতি | bālati | Water Bucket |
| 49 | W220 | বালু | bālu | Sand |
| 50 | W351 | এইডস্ | Ē'iḍasa | AIDS |
| 51 | W352 | বাত | baat | Arthritis |
| 52 | W353 | ব্যান্ডেজ | byānḍēja | Bandage |
| 53 | W354 | ক্যাপসুল | kyāpasula | Capsule |
| 54 | W355 | চিকিৎসা | cikitsā | Treatment |
| 55 | W356 | চোখ ওঠা | cōkha ōṭhā | Conjunctivitis |
| 56 | W357 | ডেঙ্গু | dēṅagu | Dengue |
| 57 | W358 | ডক্টর | dakṭara | Doctor |
| 58 | W359 | দংশন | danśana | Bite |
| 59 | W360 | দুর্বল | durbala | Weak |

During the sessions, participants were seated, resulting in the non-active hand consistently being outside the video frame. Variations in camera position led to depth variation and background variations. Sometimes volunteers were in angled sitting positions and were not always centered in the video frames. Different variations of Banlga sign volunteers are illustrated in Figure 1.

The sign users observed the expert and recorded the signs subsequently, which were then certified by the expert as unambiguous. The sign users repeated the signs in their natural hand dominance. Remarkably, some sign users switched their hand dominance from left-handed to right-handed and vice-versa, even in the mid-session. In BdSL silent mouth utterance is encouraged not obligatory and sign users were well-educated about it. So, many sign users skipped mouth utterances. Those who pronounced the sign word, among them some pronounced after the sign itself is finished (delayed pronunciation). All the signs were deemed valid and were certified by the sign expert.

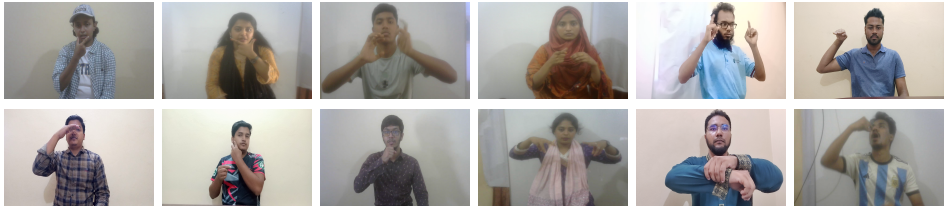

**Fig. 1**: Illustration of Variations in hand dominance, position, depth, background, and appearance.

### 3.2 Data Annotation

Sign users repeated sign words typically 10-15 times in a video. We needed to annotate the dataset to mark the user number, hand dominance/orientation, camera view (front or lateral), and frame rate and to extract frame numbers of the start/end gestures by analyzing video frames, writing word numbers, and trial numbers, and tag the video with the corresponding Bangla sign words.

There were 60 annotators assigned to annotate the dataset correctly. All of the annotators were undergraduate students of Computer Science and Engineering (age range 19-22 years) having technical knowledge of video editing tools to extract frame numbers from video. Before starting the annotation process, each annotator was briefed on the meaning of the words and demonstrated the sign words. They were instructed to extract the starting frame numbers of the word instances (frame of intention and frame of actual start) and ending frame numbers of the same instances(frame of gesture end, frame of complete withdrawal). Each of the annotators was employed to annotate videos for two sign words only and two annotators cross-verified each other's annotation quality.

A text annotation file was created for each two words and then the text file was used to generate Json files. The Json files were later used throughout the machine-learning pre-processing works. We decided the data of sign users U4 and U8 to be used as the test set for classification tasks which is described in section 4.2.1.

### 3.3 Difficulties in BdSLW60

দেবর (W11) and দুলাভাই (W12) are very similar, they only differ by a finger (Figure 2). The same is true for চাচা (W6) and দাদা/নানা (W8), differing in fingers, i.e. they are varied by fine-grained gestures only. চাচা has an intermediate step (touching the Chin), which needs hand-depth information from the face (Figure 3). আত্মীয় (W2) and চকলেট (W49) use a similar rotation of hands, differing in fingers (Figure 4). বাবা/আব্বা(W1) and দাদী/নানী (W9) are similar and only differ in hand orientation and fine finger movement (Figure 5). আনারস (W39) and এসি (W215) have similar gestures but in reverse order; however this pair have a good difference in orientation and finger gesture (Figure 6).

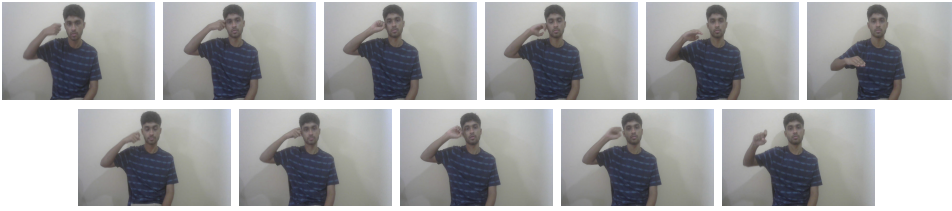

**Fig. 2**: দেবর (W11) (First row) and দুলাভাই (W12) (Second row) are very similar, only differ by a finger.

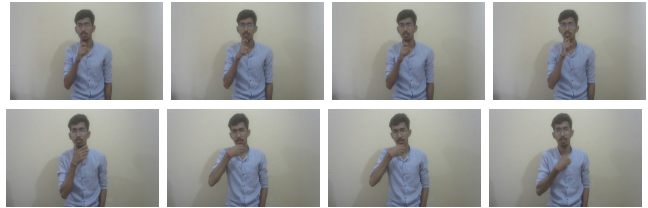

**Fig. 3**: চাচা (W6) (First row) and দাদা/নানা (W8) (Second row) are very similar, only differ by fingers. চাচা (W6) has the intermediate step of touching the chin, which needs highlighted local depth information.

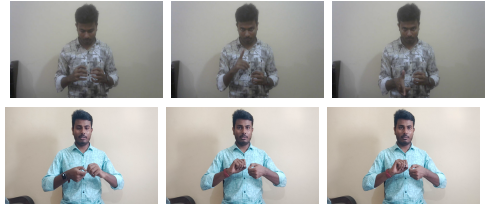

**Fig. 4**: আত্মীয় (W2) (First row) and চকলেট (W49) (Second row) given by user U10, provided a similar rotation of hands, differing in fingers only.

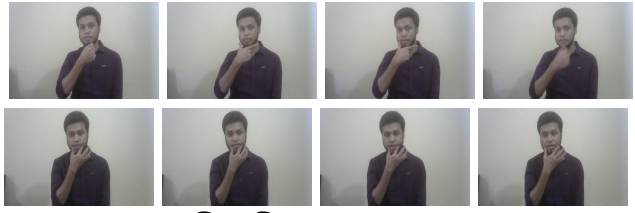

**Fig. 5**: বাবা/আব্বা (W1) and দাদী/নানী (W9) differ only by hand orientation and fine finger movement.

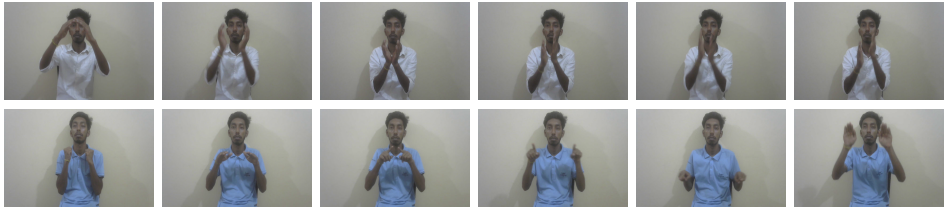

**Fig. 6**: আনারস (W39) and এসি(W215) have similar gestures but in reverse order.

### 3.4 Dataset organization

The files of the dataset have been uploaded in [dataset link] which are organized under the folders named by word numbers (e.g. W1-2). Inside each folder, there are raw video files with the naming format of user number, sign word number, and camera view (e.g. U11W1F.mp4). The annotation file for the corresponding two words is provided as a text file along with the generated JSON file containing all information needed for pre-processing.

**Table 3**: Key statistics of BdSLW60 dataset

| Characteristics | BdSLW60 |
|---|---|
| Total number of trials (Videos) | 9307 |
| Max Frame among all the trials | 164 |
| Min Frame among all the trials | 9 |
| Avg. Frame per trial | 44.20 |
| Total number of right-hand instances | 7673 |
| Total number of left-hand instances | 1634 |

## 4 Methodology

In this section, we describe our methodology of how we use our dataset for the machine learning pipeline and provide the benchmark of our proposed dataset. The overall architecture of our approach is given in Figure 7.

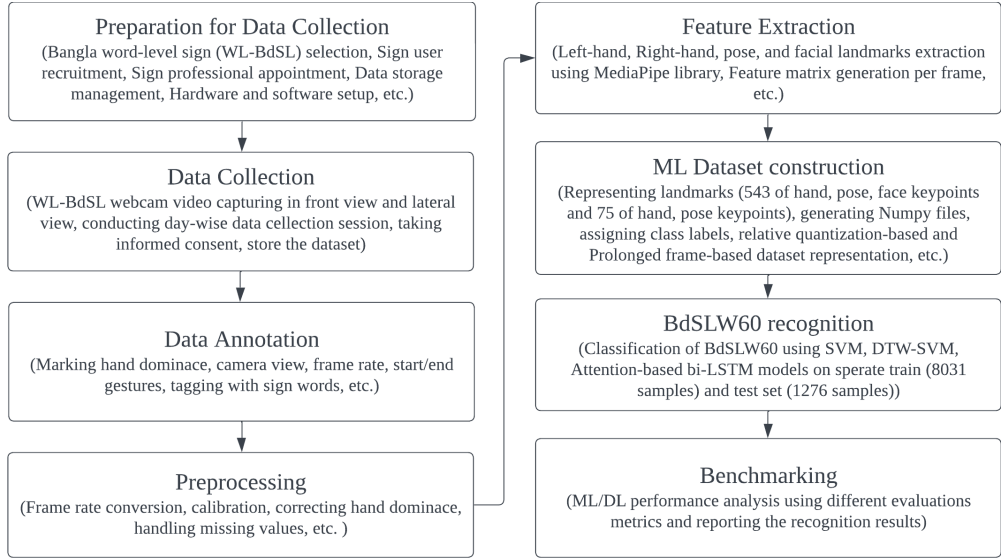

**Fig. 7**: Architecture of BdSLW60 dataset construction and recognition.

## 4.1 Preprocessing

### 4.1.1 MediaPipe Landmarks Collection

We intend to collect sign users' spatio-temporal information from left-hand, right-hand (including finger, wrist, and palm points), arms and face. Though we collected videos of the sign users, we found MediaPipe framework [33, 52] provided landmarks are used in the state-of-the-art research [11, 47–51] with better recognition accuracies. As we have manually annotated the frame numbers of start/end sign gestures, we did not go for any automated region of interest extraction or object detection techniques (e.g. YOLOv5 [53]).

We used MediaPipe Holistic [33] to extract all 543 landmarks for each frame from the video. The landmarks are three-dimensional information (x,y, and d) of MediaPipe defined keypoints. The landmarks include 33 pose keypoints, 468 face keypoints, 21 left-hand and 21 right-hand keypoints. we extracted all the frames and saved as Numpy array. The annotation JSON file and the saved Numpy arrays were used to do any sort of preprocessing work later.

### 4.1.2 Dealing with missing points

MediaPipe holistic extracts landmarks from video with a confidence score of 0.5. If any landmark cannot be extracted, it returns a NULL. In Numpy array, those are taken as zero. MediaPipe cannot estimate some hand landmarks of a rotated hand sometimes, it is already known.

Ideally, missing points should be interpolated in between frames. However, we recognized that in sign language, one hand may always be outside the video frame which will produce missing points. In our dataset, we have 41% missing hand points only. Other missing points are heap and leg points. Missing points may be part of the sign or it may be coming from MediaPipe itself. To fill the missing points, we need knowledge of hand dominance, and knowledge about the sign, which at this state of the art, is not feasible.

### 4.1.3 Calibration

MediaPipe provides normalized landmarks for a video. However, the videos we collected were taken in many sessions in different setups resulting in depth variation at least. So we re-calibrated the sign instances from the first frame of the first sign of a video. The co-ordinate center was taken as the middle point of two shoulders. This ensured uniform calibration of the depth across videos. This also corrected the non-centred sitting of the sign users.

### 4.1.4 Frame Rate Correction

There were 3 frame rates found across all videos. Though there were a few variations, we decided to annotate videos at 30fps, 24fps, and 15fps. As a higher frame rate captures better motion quality [54], we decided not to downsize the frame rates. Rather we converted the other two frame rates to 30fps by uniformly duplicating frames. So, in 15fps, every frame was duplicated, whereas, in 24fps every fourth frame was duplicated. Though interpolation or deep learning-based techniques [55] should be applied as a standard instead of duplication, we leave that as a separate research work.

### 4.1.5 Augmentation

We did not do any augmentation of the dataset as our main objective was to provide the raw BdSLW60 dataset with its intrinsic diversity and inherent challenges and to report the benchmark result. We wanted to maintain the original spatiotemporal relationships between the landmark points and the sign class labels.

### 4.1.6 Frames per Sign Instance

We gathered statistics over all the sign videos and found longest sign (including frame rate correction) has 164 frames and the minimum has only 9. Following the regular research technique, we zero-padded the remaining frames up to 164. For traditional machine learning like SVM, we also generated another version, where we prolonged the short gestures by uniformly duplicating the frames to match 164 frames. We name the dataset PROLONGED. This ensures fixed-length sequence input processing through SVM and explicitly addresses the temporal aspects of BdSLW60.

### 4.1.7 Correcting Hand Dominance

Though traditionally hand dominance variations, i.e. left-handedness or right-handedness, are not corrected expecting ML/DL classifiers to learn it, we saw a scope of work here. Correction of hand dominance will allow keyframes to be encoded unambiguously and use NLP or lexical analysis techniques to recognize signs [2, 7, 24, 32].

There has been a recent effort to detect hand dominance and articulation using MediaPipe data [56].

We converted all the left dominant instances to right dominance, by just flipping the hands (leaving the face) horizontally centering around the middle of the two shoulders. We flipped hands as they conveyed the most information in sign language. We did not flip the face as it is mostly symmetric, MediaPipe landmarks allowed the flexibility to flip only the hands, which is not possible in image-based techniques. We name this dataset version as FLIPPED. We also kept the original left dominant data intact, for comparative study.

So in terms of dataset, we have four versions: PROLONGED: where Numpy of sign frames are uniformly spread to 164. This has also the FLIPPED version. So, two versions generated here, are used for traditional classifiers as SVM. NON-PROLONGED: This version can also be FLIPPED.

As the PROLONGED version is only used in SVM to get benchmark accuracy, we mention only PROLONGED and FLIPPED to point to four versions and it is understood from the context.

### 4.1.8 Relative and Quantization of the landmark points

We made a fifth dataset, by quantizing the landmark points (x, y, d coordinates), mainly to see the effect of the reduction of depth variations. Our previous research indicated the approach produces 3-5% increased accuracy on dynamic gestures such as air writing [27, 29].

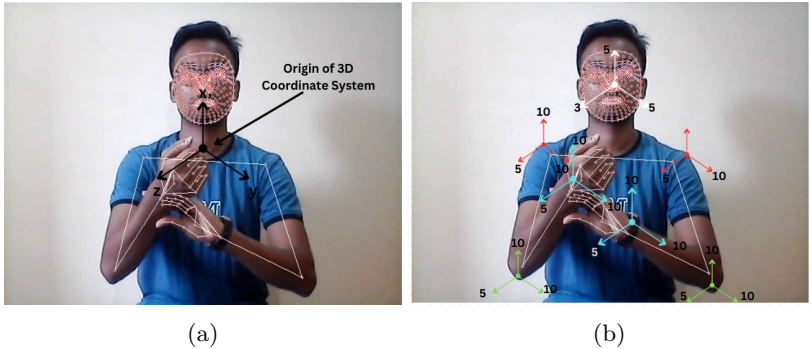

(a) (b)

**Fig. 8**: (a) Calibrated frame of reference (b)Transformed frame of references with Relative Quantization levels

In this paper, we do a bit more; we first translated the global coordinates to local coordinates following their natural physiological centers. We transferred the center of the hand points to wrists, the center of the face points to the nose, the center of the elbow points to shoulders, the center of the ankle points to heaps, and so on. Shoulder and heap points were unchanged and carried the global coordinates from the calibration step. We did the coordinate translation to highlight the local change of limbs, especially hand points. We then quantized x, y, d to a fixed number of levels. We allowed hand points to have (10, 10, 5) quantization levels and face points to have (5, 5, 3) levels. Heap and leg points were fixed to one single level meaning ignoring their impact. So, effectively through the calibration and then the relative coordinate translation with quantization, each frame's landmark information can be processed locally except the the two shoulder points are processed with the global calibration point. It would be interesting to study the effect of body movement on the accuracy using the globally calibrated shoulder points.

In this paper, we used only pose and hand landmarks from this dataset. However, pose landmarks also include 10 face points. heap and leg points are also included within pose landmarks [57]. We name this dataset as RQ.

### 4.1.9 Normalization

Normalization is essential for both traditional classifiers and deep learning-based models. While deep learning models may not require traditional normalization, the inclusion of batch normalization is essential to equalize the impact of depth variations. Calibration reduces the variation in different camera settings, normalization ensures the scale variation in different video conditions.

## 4.2 Experiments and Result analysis

To provide the benchmark of our dataset, we have experimented with the validity of our dataset in two approaches, using the classical ML models and using DL model.

### 4.2.1 Training and Testing instances

We have 18 sign users and a total of 9307 sign instances. We reserved signs of two sign users (U4 and U8) for the test set and used others in the training set (Table 4). If we make a train-test split based on sign instances blindly, signs of the same user obviously will go to both the training and test set and will bias test accuracy.

### 4.2.2 Result Metrics

For our proposed BdSLW60 dataset, we employ randomized 10-fold cross-validation on the training set, ensuring robust evaluation and biases. We report the average cross-validation accuracy and prioritize the best test accuracy by comparing the test data with each fold model, addressing variations across subsets. We emphasize true positive (top-1) results as it reflect the model's precision in identifying the specific gestures associated with BdSL words.

### 4.2.3 Feature sets and Experiments

We have experimented with two different feature sets: (i) all 543 MediaPipe landmarks points or $543 \times 3 = 1629$ features and (ii) MediaPipe landmarks excluding the face, meaning $75 \times 3 = 225$ features. The motivation for defining these two feature sets was to conduct a comparative analysis of the impact of facial expressions, including silent mouth utterances, on recognition accuracy.

Table 4: Training and testing instances for classification of BdSLW60 dataset

|  | Sign User | RH[1] Instances | LH[2] Instances | Total |
|---|---|---|---|---|
| Test | U4 | 532 | 105 | 637 |
|  | U8 | 611 | 28 | 639 |
|  | Total testing instances | 1143 | 133 | 1276 |
| Train | U1 | 321 | 654 | 975 |
|  | U2 | 585 | 159 | 744 |
|  | U3 | 583 |  | 583 |
|  | U5 | 571 | 29 | 600 |
|  | U6 | 784 | 31 | 815 |
|  | U7 | 107 |  | 107 |
|  | U9 | 646 | 10 | 656 |
|  | U10 | 235 | 148 | 383 |
|  | U11 | 679 | 92 | 771 |
|  | U12 | 617 | 38 | 655 |
|  | U13 | 435 | 186 | 621 |
|  | U14 | 112 |  | 112 |
|  | U15 | 633 | 22 | 655 |
|  | U16 | 57 | 40 | 97 |
|  | U17 | 24 | 92 | 116 |
|  | U18 | 141 |  | 141 |
|  | Total training instances | 6530 | 1501 | 8031 |

[1] Right Hand
[2] Left Hand

Though pose landmarks in the second feature set inherently contain 10 face points, we chose not to exclude them to maintain consistency and avoid feature handcrafting in this study. It's important to note that, except for Support Vector Machines (SVM), we consistently worked with the NON-PROLONGED dataset. Additionally, we worked with a FLIPPED or ORIGINAL (non-flipped) version of the dataset and experimented with two different feature sets.

### 4.2.4 Using Classical ML models

#### *SVM*

SVM is a non-temporal classifier. So, we flattened all 164 frame features and fed them to SVM. This technique of converting temporal features to spatial features worked well. We also helped the classifier by minimizing the temporal variations (9-164 frames) across sign words by using the PROLONGED dataset. It is like going in slow motion to cover a time duration. The number of features for two possible experiments is (i) $1629 \times 164 = 267,156$ and (ii) $225 \times 164 = 36,900$.

With the RBF kernel, SVM did fit well with validation data but performed poorly with the test set. We got good test accuracy with linear kernel.

### *SVM with DTW*

Before LSTM was introduced for dynamic gesture recognition, SVM with DTW features was a standard technique to deal with statio-temporal features [27, 28]. The only problem with this approach is its prohibitive computation requirement, owing to the sequential dynamic programming algorithm of DTW feature calculations.

We took 60 sign word templates from the test set randomly and compared a sign instance with all 60 sign word templates and made them as features for SVM. DTW distance finds a holistic distance measure of any temporal feature but does not capture other temporal details. However, it will always provide a reasonable classification accuracy. DTW distance calculation is prohibitively expensive but we still went for it because of our lack of knowledge with deep learning techniques at the beginning. Nevertheless, we could juxtapose the accuracy of classifiers of two different era, thanks to our naiveness. The number of features for two possible experiments are (i) $1629 \times 60 = 97,740$ and (ii) $225 \times 60 = 13,500$.

**Table 5**: Test Accuracy of Traditional Classifiers

| Classifier | with all 543 Landmarks | | | with 75 Pose and Hand Landmarks | | |
|---|---|---|---|---|---|---|
| | FLIPPED | ORIGINAL | Avg CV[3] | FLIPPED | ORIGINAL | Avg CV[3] |
| SVM[1],[2] | 57.4 | 49.1 | 97.0 | 67.6 | 60.2 | 97.9 |
| SVM[1]+DTW | 41.6 | — | 98.67 | 65.8 | — | 98.9 |

DTW weights were calculated in 60 different computers in several days

[1] with linear kernel

[2] used PROLONGED dataset

[3] In FLIPPED dataset

### 4.2.5 Using Deep Learning based models

#### *Attention-based Bi-LSTM*

Our constructed BdSLW60 dataset is a spatio-temporal dataset consisting of Bangla sign word gestures. So, naturally, it rightly fits spatio-temporal deep learning-based models like LSTM, bi-LSTM, GRU, etc. To benchmark the result we chose bi-LSTM on the NON-PROLONGED version of BdSLW60 as it adapts well in large sequences merging information from backward and forward directions and embedding temporal dependencies in both directions.

Initially, we tried a model that is already published and works well with skeletal data [29]. However, when applied to our dataset BdSLW60, the model could not find any gradient with MediaPipe landmarks. We understood we had many missing points, so we set the dropout of the two layers to 0.3. It produced some 10-12% accuracy with 543 landmark features which we could further improve to 32% by scaling the normalized feature values by 100. Then we applied the Attention layer and allowed one Bi-LSTM output to go to the dense layer in the network. this perfectly found the gradient.

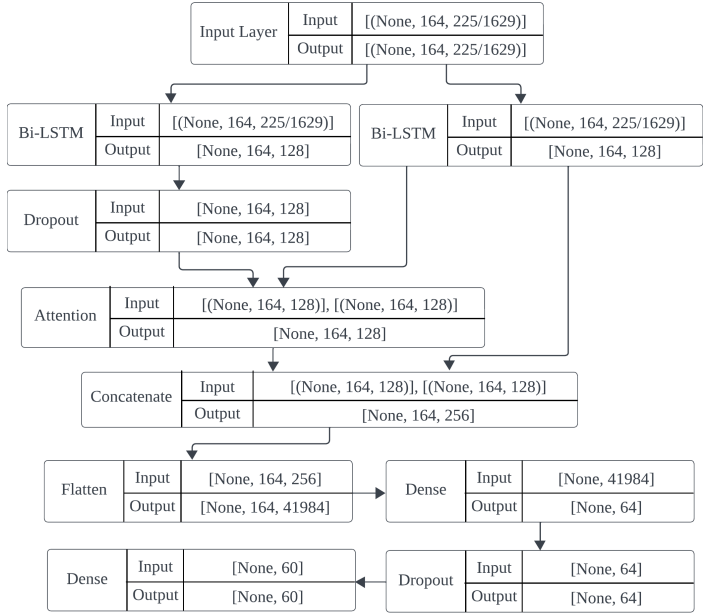

**Fig. 9**: Attention-based bi-LSTM architecture.

Figure 9 shows the architecture of our Attention-based Bi-LSTM network. We fixed the dropout rate to 0.3 and learning rate to 0.00003 after trail and error.

## 5 Results and Discussion

The results are presented in two tables: Table 5 and Table 6 respectively for traditional classifiers and our deep learning architecture. Traditional classifier SVM always provided better accuracy with flipped datasets with a benchmark accuracy of 67.6% (confusion matrix in Figure 10). This encouraged us to further experiment with hand dominance correction in deep learning setup. The SVM-DTW algorithm could not reach the accuracy of the SVM itself. After experimenting, we realized the cause that DTW, being holistic distance-based, cannot capture the temporal relationships and variations among features. We found excluding face landmark features always produces better accuracy in our dataset. This might be because, most of the sign users did not pronounce the sign word, while some did and hence introduced confusion for the classifiers.

In Attention-based Bi-LSTM we started with a batch size 32, with 200 epoches. We set early stopping conditions with patience equal to 5. However, we found the gradient is retained for datasets with 75 pose,hand landmarks and with more epochs increased the accuracy by 1-3%. We could not obtain a conclusive result for the hypothesis "hand dominance correction produces better accuracy" in the DL setup. We felt that the DL network having more parameters could generalize to capture the variation of the dataset, while the SVM could not. With 75 landmarks, we increased batch sizes

**Table 6**: Test Accuracy of Bi-LSTM with Attention on NON-PROLONGED BdSLW60 dataset

| Batch, Epoch[1] | with all 543 Landmarks | | | wtih 75 Pose and Hand Landmarks | | |
|---|---|---|---|---|---|---|
| | FLIPPED | ORIGINAL | Avg CV[2] | FLIPPED | ORIGINAL | Avg CV[2] |
| 32, 200(p=5) | 58.9 | 49.7 | 90.2 | 64.1 | 61.3 | 96.6 |
| 32, 200 | 46.1 | | 86.2 | | | |
| 64,200(5) | 45.6 | | 88.2 | 64.3 | 63.3 | 96.8 |
| 64,200 | | | | 65.4 | 64.5 | 97.3 |
| 72,200 | | | | 63.8 | | 97.6 |
| 64,200(5) RQ[3] | | | | 72.5 | | 94.0 |
| 64,200, RQ[3] | | | | 75.1 | | 95.4 |

[1]Early stopping patience given in bracket

[2]In FLIPPED dataset

[3]Used RQ dataset

64 and 72 to see which one is best suited. We found that 64 batch size give the best accuracy for the dataset with just pose and hand landmarks. This is intuitive as the number of parameters is much less compared to its counterpart(225 features are 7.24 times smaller than 1629), it needed a bit of overfitting. The relative Quantization technique produced much better test accuracy up to 75.1% which we report as the benchmark accuracy for our dataset BdSLW60. The cross-validation accuracy reached the highest of 98.9% in traditional machine learning models and 97.6% in DL model, expressing the capacity of both types of models to fit the dataset well. The confusion matrix of benchmark test accuracy of 75.1% is shown in Figure 11.

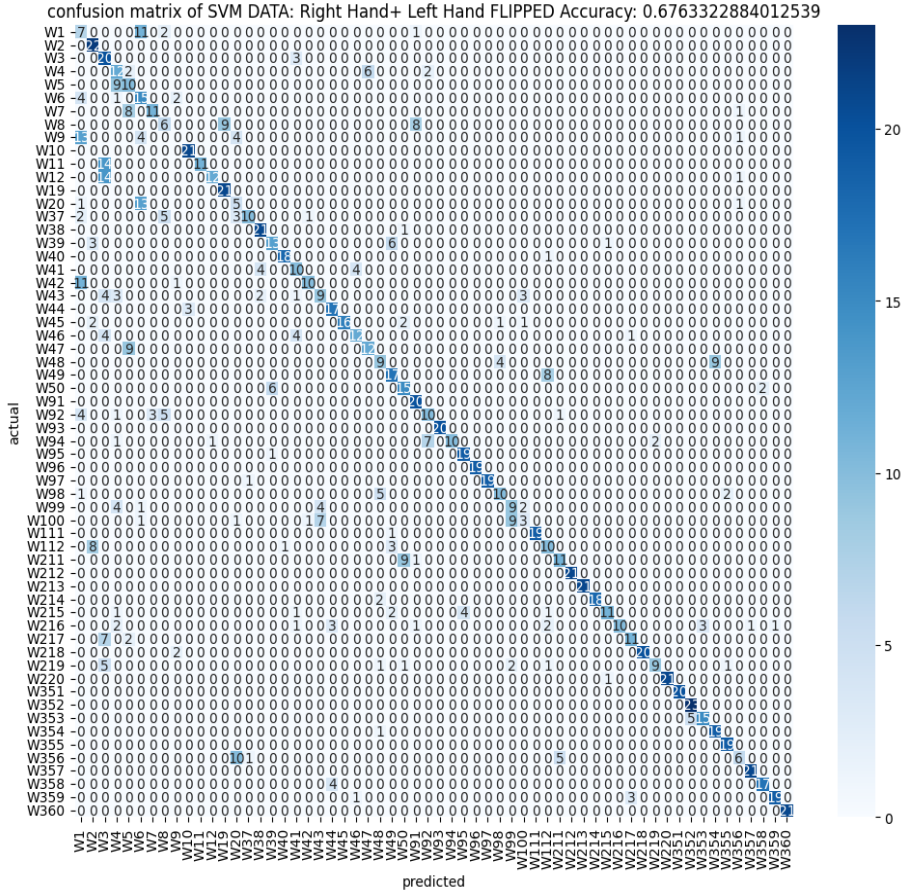

**Fig. 10**: Confusion Matrix of SVM with 75 Pose and Hand Landmarks

# 6 Conclusion and future works

Our experiments have been tailored to task-specific, i.e. increasing recognition accuracy of the BdSLW60. However, the ultimate objective is to work on continuous sign language recognition which necessitates frame-level recognition within a Natural Language Processing (NLP). The way we encoded the frames is a very good candidate for the keyframe encoding technique. We propose that our frame encoding method is well-suited for keyframe encoding techniques, offering a more efficient alternative to the memory-intensive one-hot encoding used in NLP models. Our Relative Quantization (RQ) technique demonstrates the ability to encode extensive keyframe vocabularies.

We report the benchmark result on our dataset BdSLW60 of 75.1% with just one attention layer in the deep learning network. The result could be improved with multi-head attention and with several layers of attention. Of course, diverse augmentations can also be applied which should increase the recognition accuracy further. In the future, we plan to release a larger dataset of 401 words and work on transformer-based language models. Additionally, our future efforts will include the annotation of sentence-level datasets to support end-to-end translation tasks, aligning with our broader goal of advancing continuous sign language recognition and application in real-world scenarios.

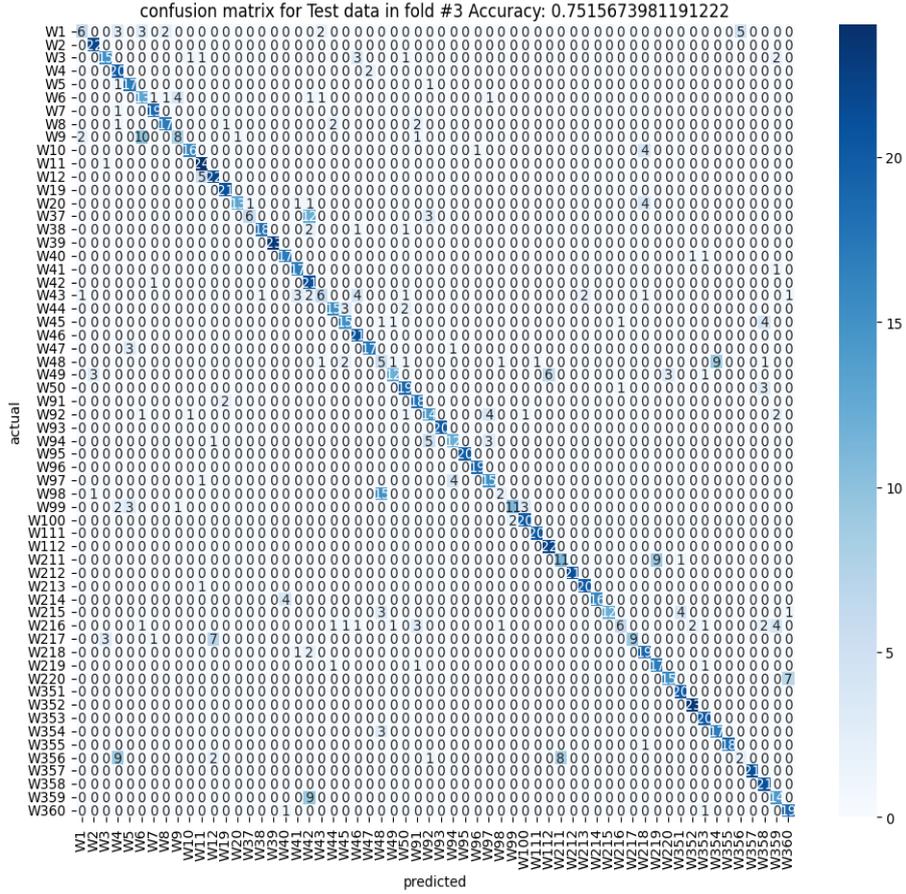

**Fig. 11**: Confusion matrix for Relative Quantization based technique